\documentclass[10pt,twocolumn,letterpaper]{article}

\usepackage[accsupp]{axessibility}  
\usepackage{cvpr}              

\usepackage{dblfloatfix}
\usepackage{times}  
\usepackage{helvet}  
\usepackage{courier}  
\usepackage[hyphens]{url}  
\usepackage{graphicx} 
\usepackage{natbib}  
\usepackage{caption} 
\usepackage{makecell}
\usepackage{algorithm}
\usepackage{tabularx}  
\usepackage{diagbox}
\usepackage{adjustbox}
\newcommand{\daash}{\textbf{\texttt{DASH}}}
\newcommand\shortsection[1]{\vspace{6pt}{\noindent\bf #1.}}
\usepackage{amsmath}
\newcommand{\bx}{\boldsymbol{x}}
\newcommand{\bxa}{\boldsymbol{x}_a}

\usepackage{newfloat}
\usepackage{listings}
\DeclareCaptionStyle{ruled}{labelfont=normalfont,labelsep=colon,strut=off} 
\floatstyle{ruled}
\newfloat{listing}{tb}{lst}{}
\floatname{listing}{Listing}
\def\paperID{34102} 
\def\confName{CVPR}
\def\confYear{2026}

\title{\daash: A Meta-Attack Framework for Synthesizing Effective and Stealthy Adversarial Examples}

\author{
  Abdullah Al Nomaan Nafi$^{1}$, 
  Habibur Rahaman$^{2}$, 
  Zafaryab Haider$^{1}$, 
Tanzim Mahfuz$^{1}$, \\
  Fnu Suya$^{3,\dagger}$, 
  Swarup Bhunia$^{2}$, 
  Prabuddha Chakraborty$^{1,*}$
\\
  $^{1}$University of Maine,
  $^{2}$University of Florida,
  $^{3}$University of Tennessee, Knoxville \\
  {\tt\small \{abdullah.nafi, zafaryab.haider, tanzim.mahfuz, prabuddha\}@maine.edu}\\
  {\tt\small rahaman.habibur@ufl.edu \quad suya@utk.edu \quad swarup@ece.ufl.edu}
}

\usepackage{booktabs}
\usepackage{multirow}
\usepackage{amsmath}
\usepackage{amsfonts}
\usepackage{algpseudocode}
\usepackage{xcolor}
\usepackage[table]{xcolor}

\definecolor{cvprblue}{rgb}{0.21,0.49,0.74}
\usepackage[pagebackref,breaklinks,colorlinks,allcolors=cvprblue]{hyperref}

\begin{document}
\maketitle
\footnotetext{* Corresponding author \quad
$\dagger$ Co-corresponding author}

\begin{abstract}

Numerous techniques have been proposed for generating adversarial examples under strict \( \ell_p \)-norm constraints. However, such norm-bounded examples often fail to align well with human perception, and only a few methods specifically explore perceptually aligned adversarial examples. Moreover, it remains unclear whether insights from \( \ell_p \)-constrained attacks can be effectively leveraged to improve perceptual efficacy. In this paper, we introduce \daash, a differentiable meta-attack framework that generates effective and perceptually aligned adversarial examples by strategically composing existing \( \ell_p \)-based attack methods. \daash~operates in a multi-stage fashion: at each stage, it aggregates candidate adversarial examples from multiple base attacks using learned, adaptive weights and propagates the result to the next stage. A meta-loss function guides this process by jointly minimizing misclassification loss and perceptual distortion, enabling the framework to dynamically modulate the contribution of each base attack throughout the stages.
 We evaluate \daash~on adversarially trained robust models across CIFAR-10, CIFAR-100, and ImageNet while considering visual perception metrics (e.g. SSIM, FID, LPIPS) in the perturbation budget (instead of \( \ell_p \)-norm). Despite relying solely on \( \ell_p \)-constrained based methods, \daash~significantly outperforms state-of-the-art perceptual attacks such as AdvAD, achieving higher attack success rates (e.g., 20.63\% improvement) and superior visual quality, as measured by SSIM, LPIPS, and FID (improvements of $\approx11$, 0.015, and 5.7, respectively). \daash~generalizes well to unseen defenses and different white-box/black-box scenarios, making it a practical and strong baseline for evaluating robustness. Code is available at: \url{https://github.com/siege-research/DASH}

\end{abstract}


%
\section{Introduction}
Adversarial examples against deep neural networks (DNNs), which add imperceptible perturbations to clean images to induce misclassification (targeted/untargeted) \cite{goodfellow2014explaining}, have attracted significant attention from the research community. Various attack methods, including FGSM \cite{goodfellow2014explaining}, PGD \cite{madry2017towards}, and CW \cite{carlini2017towards}, as well as adaptive evaluation frameworks such as AutoAttack \cite{croce2020reliable}, have been developed to benchmark both robust and undefended models. However, the majority of these approaches prioritize imperceptibility by constraining perturbations within a small $\ell_p$-norm ball, with $\ell_2$ and $\ell_{\infty}$ norms being the most common ones. 

Yet, minimizing $\ell_p$-norm does not always yield high perceptual similarity to the human visual system \cite{zhang2018unreasonable, laidlaw2020perceptual,sharif2018suitability}, despite imperceptibility being a fundamental design goal of stealthy adversarial examples. This disconnect has motivated the adoption of perceptually-aligned similarity metrics such as Structural Similarity Index (SSIM) \cite{wang2004image}, Learned Perceptual Image Patch Similarity (LPIPS) \cite{zhang2018unreasonable}, and Fréchet Inception Distance (FID) \cite{heusel2017gans} to better assess visual quality \cite{li2024advad, chen2024diffusion}. However, attacks optimized under these perceptual metrics often suffer from reduced success rates; particularly against robust models (Table~\ref{tab: main_result}), highlighting a trade-off and leaving room for improvement.

Designing effective attacks against a wide range of models, especially hardened ones, remains non-trivial and often requires deep, manual effort, an observation consistently echoed in the literature on $\ell_p$-norm attacks \cite{carlini2017towards, athalye2018obfuscated, croce2020reliable}. This complexity intensifies when attempting to simultaneously preserve high perceptual fidelity, thus underscoring the need for systematic and automated strategies that can generalize across models while aligning better with human vision.
Given the extensive body of work on $\ell_p$-bounded attacks, including those targeting robust models, it is natural to ask: 
\begin{figure*}[t!]
    \centering
    \includegraphics[width=\textwidth]{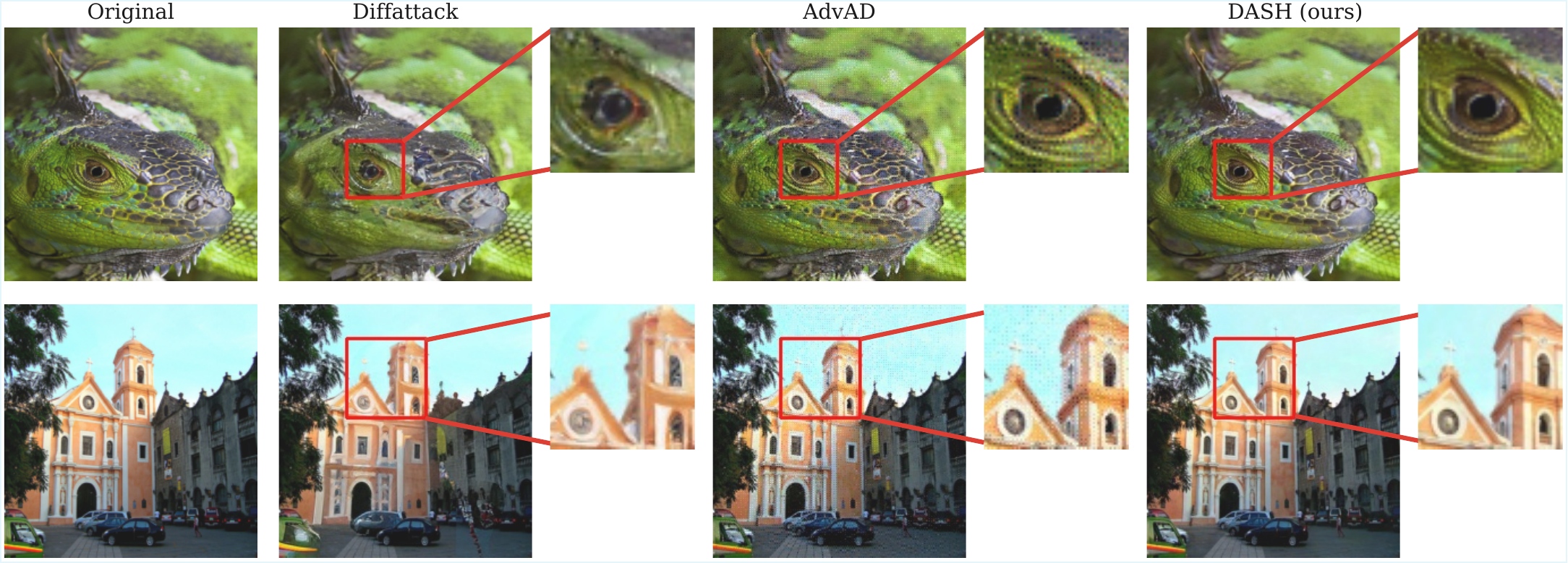}
    \caption{ Comparative visualization of adversarial perturbations across multiple attack methods (including \daash).}
    \label{fig:daash_viz}
    \vspace{-0.2in}
\end{figure*}
\emph{Can we leverage the strengths of these existing attacks to compose stronger adversarial examples that are also more perceptually aligned?}

Our key insight is that although individual $\ell_p$-norms (e.g., $\ell_2$, $\ell_\infty$) do not fully capture human perception, they each correlate with different perceptual characteristics (e.g., contrast, texture, edge disruption) \cite{sharif2018suitability, laidlaw2020perceptual}. Therefore, by combining attack strategies optimized under different norms in a principled manner, we may achieve not only improved attack success but also better perceptual quality, particularly when targeting defended models where individual attack methods may fall short.
Specifically, we propose \daash~(\textbf{D}ifferentiable \textbf{A}ttack \textbf{S}earc\textbf{H}), a fully differentiable, multi-stage framework that learns to compose multiple $\ell_p$-norm–constrained base attacks using soft attention. \daash~operates in two stages: (1) it computes a soft attention distribution over adversarial examples generated from individual candidate attacks (e.g., PGD, CW) to weight and compose the final adversarial example; (2) it iteratively refines the perturbations across multiple stages by feeding the output from one stage as the input to the next. This iterative chaining is designed to escape poor local minima inherent to the non-convex landscape of adversarial optimization.

The attention weights are optimized end-to-end through a meta-loss that jointly considers (i) attack success (via minimizing the confidence score of the ground-truth class) and (ii) perceptual similarity (via maximizing SSIM). This allows the framework to dynamically adjust the influence of each candidate attack based on the target model. Unlike existing methods that are rigid in balance between imperceptibility (from human perception) and attack success, \daash~learns to trade off these objectives adaptively across different models and datasets.

Empirically, \daash~achieves a significant improvement in both attack success rate (ASR) and perceptual fidelity compared to the current state-of-the-art. On robust models trained on CIFAR-100 such as Cui2024 \cite{cui2024decoupled}, \daash~reaches an average ASR of 99.77\% with an SSIM score of 94.43, significantly outperforming state-of-the-art perceptual attacks such as AdvAD \cite{li2024advad} (ASR: 79.14\%, SSIM: 83.18) and DiffAttack \cite{chen2024diffusion} (ASR: 77.24\%, SSIM: 91.16). Similar improvements are observed in other experimental settings. \daash~also transfers well across different models, which demonstrates its effectiveness in black-box settings.

Fig.~\ref{fig:daash_viz} presents qualitative comparisons, further illustrating the superiority of \daash~in generating perceptually plausible yet highly effective adversarial examples.

Our contributions are summarized as follows:
\begin{itemize}
\item Propose \daash, a differentiable, multi-stage meta-framework that learns to combine multiple base attacks using soft attention. \daash~decouples the objectives of attack effectiveness and stealth (e.g., SSIM), allowing for custom optimization of each using a meta-loss.
\item Validate \daash~across multiple datasets (CIFAR-10, CIFAR-100, ImageNet), seven robust models, and four post-processing defenses demonstrating substantial improvements over prior perceptual attacks.
\item Demonstrate the effectiveness of \daash~in black-box settings through rigorous transferability analysis.
\end{itemize}

\section{Background and Threat Model}

\shortsection{Adversarial Examples}
Given a deep neural network based classifier $f: \mathcal{X} \rightarrow \mathcal{Y}$, which maps inputs to class probabilities, and a clean sample $(x, y)$ where $x \in \mathcal{X}$ and $y \in \mathcal{Y}$, an untargeted adversarial example $\bxa$ is defined as:
\begin{equation}\label{eq:adv_example}
\arg\max_{i} f_i(\bx_a) \neq y \quad \text{subject to} \quad d(\bx,\bx_a) \leq \epsilon
\end{equation}
where $f_i(\bx_a)$ denotes the prediction confidence of class $i$.

\shortsection{Imperceptibility constraints} In the equation above, $d(\cdot, \cdot)$ represents a distance metric and $\epsilon$ defines the perturbation budget. $d(\cdot, \cdot)$ measures the imperceptibility of the perturbation, and the commonly used metrics include $\ell_2$ norm \cite{carlini2017towards} as well as the $\ell_{\infty}$-norm \cite{carlini2017towards}. These $\ell_p$-norm constraints may not always align with human perception, and some alternative metrics such as the Structural Similarity Index (SSIM) \cite{wang2004image} are used. SSIM is shown to better capture the luminance, contrast, and structural similarities between two images $\bx_1,\bx_2$  and is computed as below:
\begin{equation}
\text{SSIM}(\bx_1, \bx_2) = \frac{(2\mu_{x_1}\mu_{x_2} + c_1)(2\sigma_{x_1x_2} + c_2)}{(\mu_{x_1}^2 + \mu_{x_2}^2 + c_1)(\sigma_{x_1}^2 + \sigma_{x_2}^2 + c_2)}
\end{equation}
The terms $\mu_{x_1}$ and $\mu_{x_2}$ represent the mean pixel intensities of $\bx_1$ and $\bx_2$, respectively, while $\sigma_{x_1}^2$ and $\sigma_{x_2}^2$ denote their variances. The term $\sigma_{x_1x_2}$ corresponds to the covariance between $\bx_1, \bx_2$. The constants $c_1$ and $c_2$ are used to stabilize the division when the denominator is close to zero.

Besides SSIM, other perceptual metrics that better align with human perception include the Learned Perceptual Image Patch Similarity (LPIPS)~\cite{zhang2018unreasonable} and the Fréchet Inception Distance (FID)~\cite{heusel2017gans}. In this paper, we use SSIM as the perturbation constraint, it's the primary metric to optimize for imperceptibility, but we find that using LPIPS also yields similar results in practice. Using FID in the optimization process is possible but not ideal in our current experimental setup (small batch FID calculation is noisy).



\begin{figure*}[htbp]
    \centering
    \includegraphics[width=\linewidth]{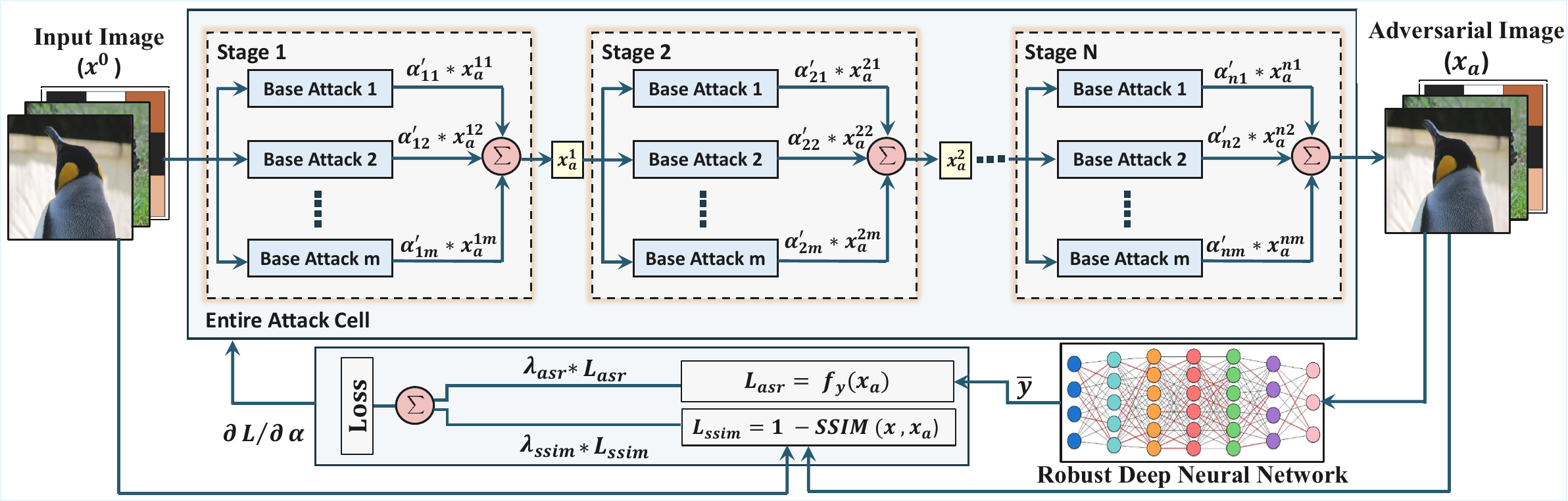}
    \caption{An overview of the proposed \textbf{D}ifferentiable \textbf{A}ttack \textbf{S}earc\textbf{H} (\daash) methodology.}  
    \label{fig:AttackCell}
    \vspace{-0.2in}
\end{figure*}




\shortsection{Threat model}
We study both white-box and black-box settings. In white-box setting, the attacker is assumed to have complete access to $f(\cdot)$, including the model architecture and parameters, which is the default for robustness analysis (e.g CW \cite{carlini2017towards}, AutoAttack \cite{croce2020reliable}) and is also considered in recent perception-aligned attacks such as DiffPGD \cite{xue2023diffusion} and AdvAD \cite{li2024advad}. In this setting, untargeted adversarial examples, as defined in Eqn.~\eqref{eq:adv_example}, can be generated via backpropagation using well-defined loss functions with respect to the input $\bx$. 
For black-box setting, it is assumed that the attacker does not have any access to the target model. We consider generating the adversarial examples using a surrogate model and then its effect is measured on the target model. Note that transfer attacks are under even stricter threat model than query-based black-box attacks where the target model returns prediction labels or confidence scores. In curiosity, we also evaluate under query-based settings, but defer the results to the supplementary materials.  


\section{Related Work}

Classical approaches for generating adversarial examples include FGSM~\cite{goodfellow2014explaining}, I-FGSM~\cite{kurakin2016adversarial}, and PGD~\cite{madry2017towards}, all of which strictly enforce a perturbation budget $\epsilon$ measured in the $\ell_{\infty}$-norm. In contrast, the CW attack~\cite{carlini2017towards} softly enforces an $\ell_2$-norm perturbation constraint by incorporating the perturbation magnitude directly into the optimization objective. Closer to our approach are AutoAttack~\cite{croce2020reliable} and Composite Attacks~\cite{mao2021composite}. AutoAttack employs an ensemble of a few selected attack methods, while composite attacks combine multiple attacks using genetic algorithms. However, these approaches still operate under conventional $\ell_p$-norm constraints and rely on simplistic or heuristic strategies for combining base attacks. In contrast, our proposed method \daash~shifts from the discrete selection strategies of prior work to a continuous and more effective interpolation framework to improve attack effectiveness and stealth.


Given the limits of norm-constrained attacks in modeling human perception, recent works shifted towards leveraging perceptual similarity metrics to guide adversarial example generation. For instance, SSAH~\cite{luo2022frequency} operates in the frequency domain to introduce perturbations in less perceptible high-frequency components, while PerC-AL~\cite{zhao2020towards} optimizes perceptual color distance, though it relies on manual hyperparameter tuning. More recent approaches exploit diffusion models by injecting perturbations into their embedding spaces, thereby influencing the denoising trajectory to produce adversarial outputs. Representative examples include DiffAttack~\cite{chen2024diffusion}, AdvAD~\cite{li2024advad}, and DiffPGD~\cite{xue2023diffusion}. 
However, these perceptual attacks are often designed as standalone strategies with limited adaptability, making them less effective across diverse target models. As shown in Table~\ref{tab: main_result}, such methods frequently underperform in both attack success and imperceptibility. By contrast, \daash~builds on the extensive literature of norm-constrained attacks, known for their compatibility with a wide range of target models, and combines them in a unified framework that enhances the effectiveness and stealth of the resulting adversarial examples. \daash~efficiently searches the convex hull of trajectories, constructs perturbations that exist between standard paths, and empirically dominates the baselines.

\section{\daash~Attack Design}
Next, we describe the high-level motivation behind the design of \daash~and then introduce the details more formally.



\subsection{Design motivation}  
Our attack design is motivated by: (1) Norm-constrained attacks sometimes exhibit complementary strengths and weaknesses \cite{croce2020reliable}. So, combining them can yield better performance, especially in under-explored settings of generating adversarial examples beyond $\ell_p$-norm constraints; (2) Adversarial example generation is a non-convex optimization process and is prone to poor local optima \cite{madry2017towards}.

Regarding Observation 1, the most natural way to combine different attacks is to take a weighted average of the adversarial examples generated by each individual attack. 
While a linear combination is practical, it introduces two key challenges: (a) How to determine optimal weights for each base attack, and (b) How to ensure that the resulting adversarial example better aligns with human perception rather than just staying within the $\ell_p$ constraints.

To address these, we make the combination weights continuous and learnable.
We introduce a meta-loss that jointly optimizes attack effectiveness and perceptual similarity to human vision. This allows adjustable weights through gradient-based optimization, effectively converting norm-constrained base attacks into perceptually aligned attacks, without needing to modify their core mechanisms. Compared to base attacks or their heuristic ensembles \cite{mao2021composite,croce2020reliable} that search over the large pixel space, we instead optimize over a much more tractable low-dimensional weight space.

Regarding Observation 2, we mitigate the risk of poor local optima by introducing a multi-stage chaining process. In each stage, we recombine base attacks using newly learned weights and feed the output adversarial example from the previous stage into the next as input. This design allows the attack to iteratively refine the adversarial perturbation. Prior work on transfer-based attacks has shown the benefits of chaining attacks across stages \cite{suya2020hybrid}, but typically only use two stages and assign different attacks to each. In contrast, our method allows arbitrary depth and continuously optimized attack combinations across all stages.

\begin{algorithm}[ht]
    \caption{Search for Optimal Attack Sequence $\mathcal{A}_\Theta$}
    \label{algo:search_OAS}
    \begin{algorithmic}[1]
        \State \textbf{Randomly Initialize} $\boldsymbol{\alpha}$ \\ 
        $\boldsymbol{\alpha} = {(\alpha_{11},\alpha_{12},\dots,\alpha_{1M}),., (\alpha_{N1},\alpha_{N2},\dots,\alpha_{NM})}$
        \For{$\text{epoch}=1$ \textbf{to} $T$}
            \State Sample a mini‑batch $\{(\bx,y)\}\subset \mathcal{D}$
            \State Generate adversarial examples: $\bx_a \gets \mathcal{A}(\bx,y)$
            \State Feed to target model: $f(\bx_a)$
            \State Compute total loss $\mathcal{L}_{\text{total}}$
            \State Update parameters: 
                  $\boldsymbol{\alpha} \gets 
                  \boldsymbol{\alpha} - \eta\,\nabla_{\boldsymbol{\alpha}}
                  \mathcal{L}_{\text{total}}(\bx_a,y)$
        \EndFor
        \State \textbf{Return} attack sequence $\mathcal{A}_\Theta$ that minimizes $\mathcal{L}_{\text{total}}$
    \end{algorithmic}
\end{algorithm}

\subsection{Design Details}
Next, we introduce our attack design more formally. We assume a total of $M$ base attacks and $N$ stages (or attack cells), both of which are hyperparameters that the attacker can freely choose to optimize attack performance. The set of base attacks can also be easily expanded as new attack methods become available in the future.

In a given stage $j$, we denote the learnable weight for each base attack as $\alpha_{j,i}$, where $i$ indexes the base attack. The output of the $j$-th attack cell is then defined as:
\begin{equation}\label{eq:attack_cell}
    \mathcal{A}_j(\bx^{j-1}_a, y) = \sum_{i=1}^{M} \frac{\exp(\alpha_{j,i})}{\sum_{k=1}^{M} \exp(\alpha_{j,k})} \cdot \bx_a^{j,i}
\end{equation}
where $\bx_a^{j,i}$ denotes the adversarial example generated by the $i$-th base attack at stage $j$. 
This linear formulation, with softmax probabilities providing normalized importance weights, provides a simple and extensible mechanism for combining base attacks while preserving the overall perceptual similarities. Future work may explore more advanced combination strategies beyond weighted averaging, or treat individual hyperparameters of base attacks, which are fixed in this work, as additional learnable variables.

After the adversarial example generation in each stage, we apply multi-stage chaining as follows: $\bxa = \mathcal{A}_N(\bxa^{N-1}, y) \circ \cdots \circ \mathcal{A}_1(\bx^0, y)$
where the composition operator $\circ$ indicates the output of each stage is passed as input to the next. This enables iterative refinement of the adversarial perturbation over multiple stages. While we adopt a linear chaining strategy here, other chaining mechanisms (e.g., cyclic or attention) are possible and left as future work.

Finally, we define the meta-loss that guides the optimization of the attack weights:
$
\mathcal{L}_{\text{total}} = \lambda_{\text{asr}} \cdot f_y(\bxa) + \lambda_{\text{ssim}} \cdot (1 - \text{SSIM}(\bx, \bxa))
$.
The first term encourages attack effectiveness by maximizing the confidence drop in the ground-truth class $y$ for the adversarial example $\bxa$. The second term penalizes perceptual distortion, measured using the Structural Similarity Index Measure (SSIM), which ranges from 0 to 1, with higher values indicating greater similarity. We include SSIM in the meta-loss to constrain \daash~adversarial examples within a visual perturbation budget. While it is not possible to strictly clip an example to a specific SSIM/FID/LPIPS budget limit, the computation of these metrics remains differentiable and hence can be used as an optimization parameter. In the experiment, the weights are optimized until some threshold SSIM is not achieved. In practice, we observe that substituting SSIM with other perceptual similarity metrics yields comparable results. Our full attack algorithm is summarized in Algorithm~\ref{algo:search_OAS} and a visual demonstration is also shown in Figure \ref{fig:AttackCell}.

\section{Experiments and Results}
Next we discuss our core experimental results. All results are obtained using NVIDIA A100 GPU (unless specified). Additional experimental results and details (e.g. black-box attacks as \daash~base attacks, \daash~attack weight distributions, base attack pool variations) can be found inside the supplementary document. 

\subsection{Experimental Setup}
We discuss the datasets, robust models, post-processing defenses, metrics, base attacks, and DASH training.
\subsubsection{Datasets and Robust Models}
Our experiments are conducted on three commonly used benchmark datasets of different scales: CIFAR-10 \cite{krizhevsky2014cifar}, CIFAR-100 \cite{Krizhevsky09learningmultiple}, and ImageNet \cite{deng2009imagenet}. For each dataset, we randomly sample 1,000 test images. Since standard models without explicit defenses are easily broken by existing attacks, our evaluation focuses on robust models curated by RobustBench \cite{croce2020robustbench}, a widely used benchmark for adversarial robustness in which the models are variants of standard adversarial training \cite{madry2017towards}.
\begin{table*}[htbp]
\centering
\caption{Performance of attacks across datasets and RobustBench \cite{croce2020robustbench} models. Base is the ASR without any post-processing defenses; 
Avg. is the average of all the ASR values. SSIM values are scaled to [0,100]. Time (sec) is the average generation time of one example.
}
\label{tab: main_result}
\resizebox{\textwidth}{!}{%
\renewcommand{\arraystretch}{1}
\small\addtolength{\tabcolsep}{-0.5pt}
\begin{tabular}{lclcccccccccc}
\toprule 
\multirow{2}{*}{\textbf{Dataset}} &
\multirow{2}{*}{\textbf{Target Model}} &
\multirow{2}{*}{\textbf{Attack}} &
\multirow{2}{*}{\textbf{Time (s)}} &
\multicolumn{6}{c}{\textbf{ASR (↑)}} &
\multirow{2}{*}{\textbf{SSIM (↑)}} &
\multirow{2}{*}{\textbf{LPIPS (↓)}} &
\multirow{2}{*}{\textbf{FID (↓)}} \\
\cline{5-10}\\[-1.8ex]
 & & & & \textbf{Base} & \textbf{JPEG} & \textbf{TVM} & \textbf{Ensemble} & \textbf{NRP} & \textbf{Avg.} & & & \\
\midrule
\multirow{26}{*}{CIFAR-100} &
\multirow{9}{*}{\shortstack{Cui2024 \\ \cite{cui2024decoupled}}}
 & DI-FGSM   & 0.17 & 65.20 & 65.49 & 65.35 & 65.55 & 57.03 & 63.72 & 92.08 & 0.0168 & 54.93 \\
& & CW	& 0.11 & \textbf{100.00} &	\underline{87.70} &	\underline{96.68} &	78.61 &	57.71 &	\underline{84.14} &	\underline{94.38} &	\underline{0.0153} &	\textbf{32.49} \\
& & PI-FGSM++ & 0.22 & 64.41 & 65.78 & 64.32 & 65.64 & 50.86 & 62.20 & 91.12 & 0.0199 & 58.94 \\
& & AutoAttack & 0.33 & \underline{82.56} & 81.33 & 79.51 & 80.55 & 59.88 & 76.77 & 92.09 & 0.0167 & 41.68 \\
& & DiffPGD   & 4.21 & 76.20 &	76.00 &	75.80 &	75.60 &	75.40 &	75.80 &	91.15 &	0.0165 &	43.50 \\
& & DiffAttack & 12.44 & 77.50 & 77.20 & 77.70 & 77.40 & \underline{76.40} & 77.24 & 91.16 & 0.0169 & 42.30 \\
& & AdvAD   & 0.51  & 82.20 & 82.80 & 81.10 & \underline{81.70} & 67.90 & 79.14 & 83.18 & 0.0288 & 45.67 \\
\cline{3-13} \\[-1.8ex]
& & \cellcolor{gray!20}{\textbf{\daash~(ours)}} & \cellcolor{gray!20}{0.96}    & \cellcolor{gray!20}{\textbf{100.00}} & \cellcolor{gray!20}{\textbf{99.90}} & \cellcolor{gray!20}{\textbf{100.00}} & \cellcolor{gray!20}{\textbf{99.32}} & \cellcolor{gray!20}{\textbf{99.61}} & \cellcolor{gray!20}{\textbf{99.77}} & \cellcolor{gray!20}{\textbf{94.43}} & \cellcolor{gray!20}{\textbf{0.0139}} & \cellcolor{gray!20}{\underline{41.08}} \\
\cline{2-13} \\[-1.8ex]
& \multirow{9}{*}{\shortstack{Wang2023 \\ \cite{wang2023better}}}
 & DI-FGSM & 0.17   & 66.46 & 67.01 & 66.07 & 66.72 & 59.86 & 65.22 & 91.96 & 0.0169 & 55.24 \\
& & CW & 0.12 &	\textbf{100.00} &	\underline{85.74} &	\underline{95.31} &	76.95 &	59.86 &	\underline{83.57} &	\underline{94.55} &	\underline{0.0161} &	\textbf{32.19} \\
& & PI-FGSM++ & 0.22 & 66.17 & 67.21 & 65.29 & 66.76 & 52.30 & 63.55 & 91.14 & 0.0199 & 57.88 \\
& & AutoAttack & 0.39& \underline{83.50} &	82.91 &	81.05 &	\underline{81.64} &	62.60 & 78.34 &	91.25 &	0.0189 &	42.19 \\
& & DiffPGD & 4.77   & 79.10 &	76.10 &	74.20 &	73.80 &	73.40 &	75.32 &	91.50 &	\underline{0.0161} &	51.80 \\
& & DiffAttack & 12.96 & 76.90 & 76.60 & 76.40 & 76.80 & \underline{76.30} & 76.53 & 90.90 & 0.0176 & 42.20 \\
& & AdvAD & 0.51     & 79.90 & 79.60 & 79.90 & 79.50 & 67.20 & 77.42 & 83.07 & 0.0293 & 47.29 \\
\cline{3-13} \\[-1.8ex]
& & \cellcolor{gray!20}{\textbf{\daash~(ours)}} & \cellcolor{gray!20}{0.79}    & \cellcolor{gray!20}{\textbf{100.00}} & \cellcolor{gray!20}{\textbf{100.00}} & \cellcolor{gray!20}{\textbf{100.00}} & \cellcolor{gray!20}{\textbf{99.12}} & \cellcolor{gray!20}{\textbf{99.22}} & \cellcolor{gray!20}{\textbf{99.67}} & \cellcolor{gray!20}{\textbf{94.73}} & \cellcolor{gray!20}{\textbf{0.0141}} & \cellcolor{gray!20}{\underline{41.59}} \\
\cline{2-13} \\[-1.8ex]
& \multirow{9}{*}{\shortstack{Addepalli2022 \\ \cite{addepalli2022efficient}}}
 & DI-FGSM & 0.18   & 72.09 & 72.13 & 71.84 & 71.74 & 63.42 & 70.24 & 92.05 & 0.0156 & 56.69 \\
& & CW & 0.13 &	\textbf{100.00} &	\underline{86.67} &	\underline{96.45} &	79.04 &	58.72 &	\underline{84.18} &	\underline{94.94} &	\textbf{0.0118} &	\textbf{27.54} \\
& & PI-FGSM++ & 0.23 & 71.50 & 71.93 & 70.86 & 71.15 & 56.89 & 68.47 & 91.14 & 0.0192 & 59.94 \\
& & AutoAttack & 0.39 & 83.77 & 82.70 & 81.02 & 80.68 & 61.33 & 77.90 & 93.03 & 0.0147 & 39.59 \\
& & DiffPGD & 4.27 &	82.50 &	82.20 &	81.80 &	81.60 &	80.90 &	81.80 &	91.05 &	0.0168 &	42.80 \\
& & DiffAttack & 12.86& 81.60 & 81.60 & 81.40 & 81.40 & \underline{81.50} & 81.35 & 91.26 & 0.0163 & 42.15 \\
& & AdvAD & 0.54     & \underline{84.50} & 84.00 & 83.70 & \underline{84.30} & 70.50 & 81.60 & 83.92 & 0.0291 & 45.63 \\
\cline{3-13} \\[-1.8ex]
& & \cellcolor{gray!20}{\textbf{\daash~(ours)}} &  \cellcolor{gray!20}{0.84}   & \cellcolor{gray!20}{\textbf{100.00}} & \cellcolor{gray!20}{\textbf{99.90}} & \cellcolor{gray!20}{\textbf{98.93}} & \cellcolor{gray!20}{\textbf{98.63}} & \cellcolor{gray!20}{\textbf{98.63}} & \cellcolor{gray!20}{\textbf{99.22}} & \cellcolor{gray!20}{\textbf{95.05}} & \cellcolor{gray!20}{\underline{0.0131}} & \cellcolor{gray!20}{\underline{39.58}} \\
\midrule
\multirow{24}{*}{CIFAR-10} &
\multirow{8}{*}{\shortstack{Bartold2024 \\ \cite{bartoldson2024adversarial}}}
& DI-FGSM & 0.72   & 41.39 & 42.46 & 42.27 & 42.75 & 35.70 & 40.91 & 90.88 & 0.0200 & 59.07 \\
& & CW & 0.62&	\underline{97.27} &	61.62 &	\underline{96.09} &	60.35 &	58.59 &	\underline{74.78} &	\textbf{95.25 }&	0.0206 &	55.25 \\
& & PI-FGSM++ & 0.83 & 45.80 & 48.46 & 46.72 & 48.89 & 23.24 & 42.62 & 88.74 & 0.0278 & 66.91 \\
& & AutoAttack & 2.17& 73.24 &	71.88 & 72.75 &	71.19 &	39.36 &	65.68 &	87.51 &	0.0264 & 53.05 \\
& & DiffPGD & 4.80 &	53.20	&52.40	&52.60 &	52.40	& 52.80	& 52.68 &	91.60 &	\underline{0.0165} &	54.20 \\
& & DiffAttack & 15.91  & 73.60 &	\underline{73.30} &	73.60 &	\underline{73.50} &	\underline{72.90} &	73.38 &	82.90 &	0.0369 &	51.60 \\
& & AdvAD & 3.99     & 61.30 & {60.70} & 60.60 & 61.20 & 41.10 & 56.98 & 83.50 & 0.0269 & \underline{49.37} \\
\cline{3-13} \\[-1.8ex]
& & \cellcolor{gray!20}{\textbf{\daash~(ours)}} & \cellcolor{gray!20}{3.25}   & \cellcolor{gray!20}{\textbf{97.62}} & \cellcolor{gray!20}{\textbf{97.32}} & \cellcolor{gray!20}{\textbf{97.62}} & \cellcolor{gray!20}{\textbf{96.13}} & \cellcolor{gray!20}{\textbf{97.32}} & \cellcolor{gray!20}{\textbf{97.20}} & \cellcolor{gray!20}{\underline{93.89}} & \cellcolor{gray!20}{\textbf{0.0132}} & \cellcolor{gray!20}{\textbf{49.17}} \\
\cline{2-13} \\[-1.8ex]
& \multirow{8}{*}{\shortstack{Cui2024 \\ \cite{cui2024decoupled}}}
& DI-FGSM & 0.16  & 51.64 & 52.71 & 52.36 & 53.01 & 45.23 & 50.99 & 90.51 & \underline{0.0206} & 62.20 \\
& & CW & 0.11 &	\textbf{99.80} &	67.68 &	\underline{95.51} &	65.33 &	65.33 &	78.73 &	\textbf{95.39} &	\underline{0.0206} &	51.38 \\
& & PI-FGSM++ & 0.21 & 55.76 & 57.09 & 56.05 & 56.84 & 31.07 & 51.36 & 88.71 & 0.0272 & 68.81 \\
& & AutoAttack & 0.41& 78.42 &	77.34 &	78.22 &	76.46 &	44.34 &	70.96 &	86.88 &	0.0273 &	56.71 \\
& & DiffPGD & 5.30 &	76.20	& 75.80 &	76.50 &	76.10 &	75.40 &	76.00 &	79.80 &	0.0350 &	60.50 \\
& & DiffAttack & 12.88& \underline{82.80} &	\underline{82.50} &	83.10 &	\underline{82.70} &	\underline{82.20} &	\underline{82.66} &	81.36 &	0.0404 & 58.24 \\
& & AdvAD & 0.54     & 66.80 & 67.20 & 66.60 & 66.20 & 45.50 & 62.46 & 83.11 & 0.0265 & \underline{50.95} \\
\cline{3-13} \\[-1.8ex]
& & \cellcolor{gray!20}{\textbf{\daash~(ours)}} & \cellcolor{gray!20}{0.42} & \cellcolor{gray!20}{\textbf{99.80}} & \cellcolor{gray!20}{\textbf{99.80}} & \cellcolor{gray!20}{\textbf{99.80}} & \cellcolor{gray!20}{\textbf{99.22}} & \cellcolor{gray!20}{\textbf{99.41}} & \cellcolor{gray!20}{\textbf{99.61}} & \cellcolor{gray!20}{\underline{93.21}} & \cellcolor{gray!20}{\textbf{0.0161}} & \cellcolor{gray!20}{\textbf{50.85}} \\
\cline{2-13} \\[-1.8ex]
& \multirow{8}{*}{\shortstack{Wang2023 \\ \cite{wang2023better}}}
& DI-FGSM & 0.17  & 51.64 & 52.13 & 52.09 & 51.84 & 44.57 & 50.45 & 90.45 & \underline{0.0203} & 63.21 \\
& & CW &0.12&	\underline{98.90} &	70.21 &	\underline{94.82} &	66.11 &	51.76 &	76.36 &	\textbf{95.25} &	0.0208 &	52.86 \\
& & PI-FGSM++ &0.22 & 55.29 & 57.11 & 54.61 & 56.27 & 29.61 & 50.58 & 88.72 & 0.0265 & 70.71 \\
& & AutoAttack & 0.41& 77.93 &	77.54 &	77.73 &	76.27 &	43.16 &	70.53 &	86.83 &	0.0276 &	56.75 \\
& & DiffPGD & 5.13 &	75.80 &	75.40 &	76.10 &	75.60 &	75.10 &	75.60 &	79.50 &	0.0345 &	60.80 \\
& & DiffAttack  & 13.49 & 82.50 & \underline{82.40} & 82.50 & \underline{82.40} & \underline{82.10} & \underline{82.38} & 80.12 & 0.0429 & 57.22 \\
& & AdvAD & 0.54     & 67.40 & 68.40 & 66.40 & 67.20 & 44.10 & 62.70 & 82.64 & 0.0271 & \underline{52.78} \\
\cline{3-13} \\[-1.8ex]
& & \cellcolor{gray!20}{\textbf{\daash~(ours)}} & \cellcolor{gray!20}{0.42}   & \cellcolor{gray!20}{\textbf{98.93}} & \cellcolor{gray!20}{\textbf{98.83}} & \cellcolor{gray!20}{\textbf{98.93}} & \cellcolor{gray!20}{\textbf{98.34}} & \cellcolor{gray!20}{\textbf{98.54}} & \cellcolor{gray!20}{\textbf{98.71}} & \cellcolor{gray!20}{\underline{92.91}} & \cellcolor{gray!20}{\textbf{0.0161}} & \cellcolor{gray!20}{\textbf{52.66}} \\
\midrule
\multirow{8}{*}{ImageNet-1k} &
\multirow{8}{*}{\shortstack{Salman2020 \\ \cite{salman2020adversarially}}}
 & DI-FGSM & 1.22   & 83.90 & 83.60 & 83.00 & 82.60 & 75.40 & 81.70 & 91.42 & 0.0849 & 49.39 \\
& & CW & 0.58 &	\underline{98.50} &	88.40 &	\underline{98.20} &	76.00 &	63.70 &	85.02 &	\textbf{97.75} &	\textbf{0.0263} &	\textbf{6.77} \\
& & PI-FGSM++ & 1.75 & 87.10 & 86.80 & 85.60 & 85.10 & 74.50 & 83.82 & 90.54 & 0.0960 & 49.57 \\
& & AutoAttack & 2.22 & 94.90 & \underline{94.30} & 91.30 & \underline{90.30} & 72.60 & \underline{91.56} & 92.90 & 0.0757 & 21.78 \\
& & DiffPGD & 10.40 & 76.20 & 69.40 & 74.80 &  68.50 & 72.30 & 72.24 & 87.50 & 0.0892 &  45.60\\ 
& & DiffAttack & 20.33 & 92.50 & 86.80 & 88.90 & 87.90 & \underline{89.80} & 89.18 & 89.17 & 0.0763 & 40.78 \\
& & AdvAD & 5.61     & 89.20 & 87.90 & 87.70 & 87.90 & 86.80 & 87.90 & 71.30 & 0.1949 & 50.14 \\
\cline{3-13} \\[-1.8ex]
 & & \cellcolor{gray!20}{\textbf{\daash~(ours)}} & \cellcolor{gray!20}{5.35}   & \cellcolor{gray!20}{\textbf{98.80}} & \cellcolor{gray!20}{\textbf{97.50}} & \cellcolor{gray!20}{\textbf{98.80}} & \cellcolor{gray!20}{\textbf{93.20}} & \cellcolor{gray!20}{\textbf{96.60}} & \cellcolor{gray!20}{\textbf{96.98}} & \cellcolor{gray!20}{\underline{96.52}} & \cellcolor{gray!20}{\underline{0.0430}} & \cellcolor{gray!20}{\underline{18.64}} \\
\bottomrule

\end{tabular}
}%
\end{table*}

\subsubsection{Post-Processing Defenses}
To further test the limits of \daash~and other SOTA attacks in terms of generating potent adversarial examples, we also experiment with post-processing defenses (on top of using RobustBench models).
The post-processing defenses include JPEG compression \cite{guo2017countering}, Total Variation Minimization (TVM) \cite{wang2018adversarial}, Neural Representation Purifier (NRP) \cite{naseer2020self}, and an ensemble of four methods: bit-depth reduction \cite{xu2017feature}, JPEG, TVM, and Non-Local Means Denoising (NLM) \cite{buades2011non}.

\begin{table}[t]
\centering
\setlength{\tabcolsep}{2pt}
\renewcommand{\arraystretch}{1.1}
\small\caption{Transferability Comparison on CIFAR-100 Dataset. S represents Surrogate models and T represents Target models.}
\label{tab:transfer-DASH}
\resizebox{\linewidth}{!}{
\begin{tabular}{c l c c c | c c }
\toprule
\multicolumn{1}{c}{\diagbox{\textbf{S (\textdownarrow)}}{\textbf{T (\textrightarrow)}}} & \multicolumn{1}{l}{\textbf{Attack}} &
\multicolumn{1}{c}{\textbf{\cite{cui2024decoupled}}} &
\multicolumn{1}{c}{\textbf{\cite{wang2023better}}} &
\multicolumn{1}{c}{\textbf{\cite{addepalli2022efficient}}} &
\multicolumn{1}{c}{\textbf{Avg.(\textuparrow)}} &
\multicolumn{1}{c}{\textbf{SSIM (\textuparrow)}} \\
\midrule
\multirow{6}{*}{\textbf{\shortstack{\cite{cui2024decoupled}}}}
& PI-FGSM++    & \cellcolor{gray!15}62.20 & 59.25 & 60.01 & 59.63 & 91.12 \\
& CW         & \cellcolor{gray!15}84.14 & 51.39 & 56.48 & 53.94 & \underline{94.38} \\
& AutoAttack & \cellcolor{gray!15}76.77 & 65.98 & 64.26 & 65.12 & 92.09 \\
& DiffAttack & \cellcolor{gray!15}77.24 & 74.76 & 74.72 & \underline{74.74} & 91.16 \\
& AdvAD      & \cellcolor{gray!15}79.14 & 73.84 & 72.70 & 73.27 & 83.18 \\
& \daash      & \cellcolor{gray!15}99.77 & 89.55 & 79.69 & \textbf{84.62} & \textbf{94.43} \\
\midrule
\multirow{6}{*}{\textbf{\shortstack{\cite{wang2023better}}}}
& PI-FGSM++    & 59.96 & \cellcolor{gray!15}63.55 & 61.43 & 60.70 & 91.14 \\
& CW         & 52.73 & \cellcolor{gray!15}83.57 & 56.78 & 54.76 & \underline{94.55} \\
& AutoAttack & 63.98 & \cellcolor{gray!15}78.34 & 64.26 & 64.12 & 91.25 \\
& DiffAttack & 75.38 & \cellcolor{gray!15}76.53 & 74.70 & \underline{75.04} & 90.90 \\
& AdvAD      & 72.98 & \cellcolor{gray!15}77.42 & 72.24 & 72.61 & 83.07 \\
& \daash      & 94.53 & \cellcolor{gray!15}99.67 & 80.64 & \textbf{87.59} & \textbf{94.73} \\
\midrule
\multirow{6}{*}{\textbf{\shortstack{\cite{addepalli2022efficient}}}}
& PI-FGSM++    & 53.24 & 53.77 & \cellcolor{gray!15}68.47 & 53.51 & 91.14 \\
& CW         & 37.32 & 37.73 & \cellcolor{gray!15}84.18 & 37.53 & \underline{94.94} \\
& AutoAttack & 52.13 & 54.08 & \cellcolor{gray!15}77.90 & 53.11 & 93.03 \\
& DiffAttack & 72.44 & 54.30 & \cellcolor{gray!15}81.35 & \underline{63.37} & 91.26 \\
& AdvAD      & 64.58 & 64.82 & \cellcolor{gray!15}81.60 & \textbf{64.70} & 83.92 \\
& \daash      & 63.50 & 63.09 & \cellcolor{gray!15}99.22 & 63.30 & \textbf{95.05} \\
\bottomrule
\end{tabular}
}
\vspace{-0.15in}
\end{table}

\subsubsection{Evaluation Metrics}
We use attack success rate (ASR) to measure attack potency, and for determining human visual imperceptibility, we use: SSIM, LPIPS, and FID. 


\begin{figure}[htbp]
    \centering
    \includegraphics[width=\columnwidth]{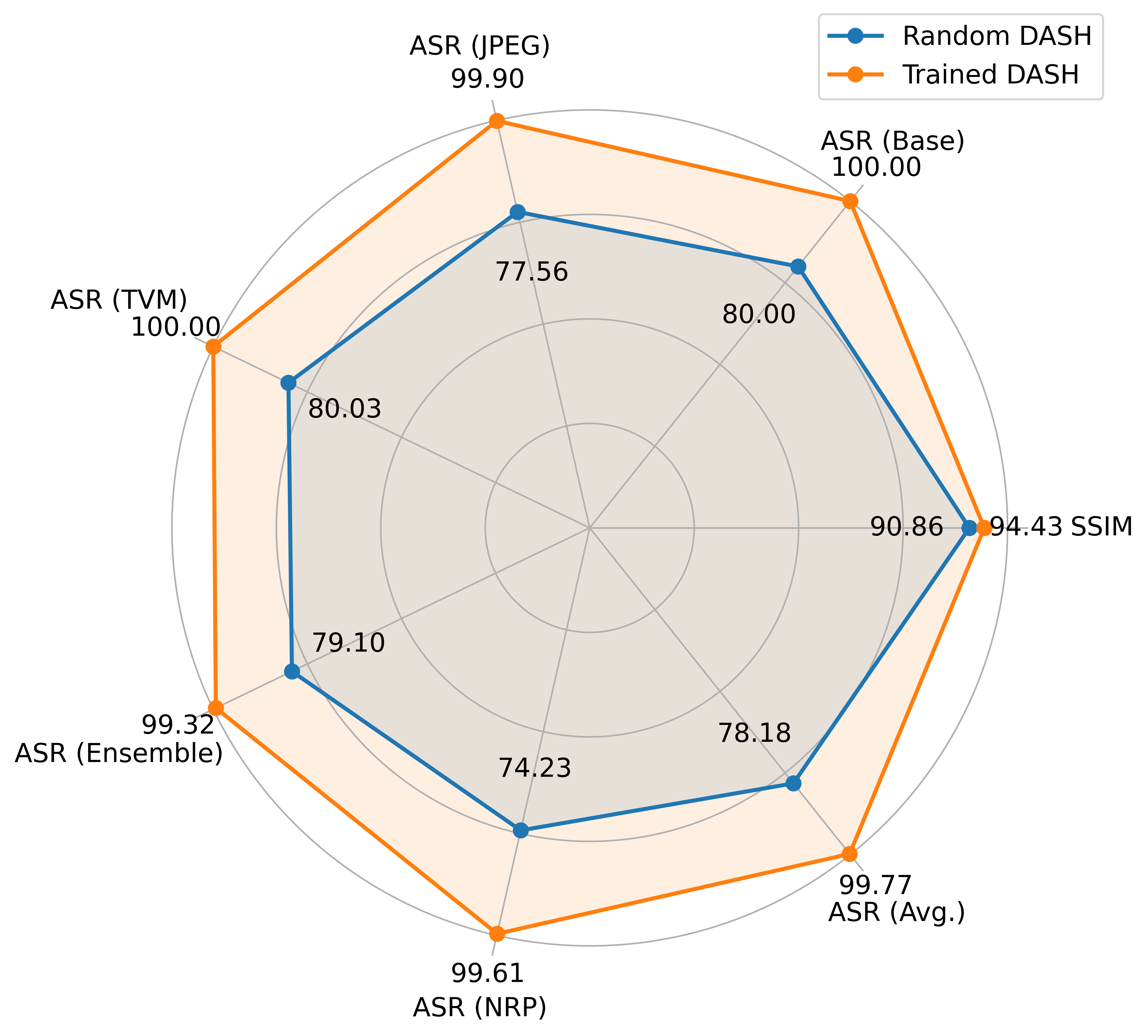}
    \caption{Comparison of \daash~performance before and after training, illustrating ASR/SSIM across different defense settings. ``Trained \daash" outperforms the randomly weighted version. 
    }
    \label{fig:random_vs_trained}
    \vspace{-0.2in}
\end{figure}
\subsubsection{Base Attacks and \daash~Training Details}
We include 10 diverse base attacks in our attack pool: FGSM \cite{goodfellow2014explaining}, PGD \cite{madry2017towards}, CW \cite{carlini2017towards}, FAB \cite{croce2020minimally}, TI-FGSM \cite{dong2019evading}, NI-FGSM \cite{lin2019nesterov}, MI-FGSM \cite{dong2017boosting}, DI-FGSM \cite{xie2019improving}, BIM \cite{kurakin2016adversarial}, and PI-FGSM++ \cite{gao2020patch}. Each attack offers unique advantages and covers a broad range of perturbation strategies. 
For $\ell_{\infty}$-norm attacks, we set the perturbation budget to $\epsilon = \frac{7}{255}$. For different variants of FGSM evaluated in this paper, we use a momentum factor of 1. In TI-FGSM~\cite{dong2019evading} and DI-FGSM~\cite{xie2019improving}, we set the \texttt{resize\_rate} to 0.9, \texttt{diversity\_prob} to 0.7, and \texttt{random\_start} to \texttt{False}. In PI-FGSM++~\cite{gao2020patch}, we set the probability of using diverse inputs to 0.7 and the \texttt{project\_factor} to 0.8.
For the Fast Adaptive Boundary (FAB) attack~\cite{croce2020minimally}, we set $\alpha_{\text{max}} = 0.1$, overshooting factor $\eta = 1.05$, and the backward step parameter $\beta = 0.9$. For BIM~\cite{kurakin2016adversarial} and PGD~\cite{madry2017towards}, we use a step size of $\alpha = 0.01$ and number of iterations $T = 10$.
For the CW attack~\cite{carlini2017towards}, we set the confidence parameter $c = 10$, margin parameter $\kappa = 0$, and optimized for 100 steps using a learning rate of $0.01$.
In addition, we include a \texttt{None} operation at each stage, which returns the input image, allowing the model to flexibly decide whether to skip processing in a given stage.
\begin{table*}[htbp]
\centering
\caption{Effectiveness of \daash~with varying numbers of stages for the robust CIFAR-100 Cui2024 \cite{cui2024decoupled} model.}
\label{tab:daash_variant}
\renewcommand{\arraystretch}{1}
\small\addtolength{\tabcolsep}{4.5pt}
\begin{tabular}{cccccccccc}
\toprule 
\textbf{\shortstack{Num. of}} & \multicolumn{6}{c}{\textbf{ASR (↑)}} & 
\multirow{2}{*}{\textbf{SSIM (↑)}} & \multirow{2}{*}{\textbf{LPIPS (↓)}} & 
\multirow{2}{*}{\textbf{FID (↓)}} \\
\cmidrule(lr){2-7}
\textbf{Stages}&  Base & JPEG & TVM & Ensemble & NRP & Avg. & & & \\
\midrule
 1 & 72.85 & 34.08 & 68.46 & 37.70 & 46.29 & 51.88 & 98.48 & 0.0027 & 16.19 \\
2 & 100.00 & 99.90 & 100.00 & 98.54 & 99.02 & 99.49 & 94.59 & 0.0115 & 43.28 \\
3 & 100.00 & 99.90 & 100.00 & 99.32 & 99.61 & 99.77 & 94.43 & 0.0139 & 41.08 \\
4 & 100.00 & 100.00 & 100.00 & 99.61 & 99.71 & 99.86 & 93.22 & 0.0190 & 47.56 \\
5 & 100.00 & 100.00 & 100.00 & 100.00 & 99.90 & 99.98 & 92.92 & 0.0209 & 48.25 \\
\bottomrule
\end{tabular}
\end{table*}

For the meta-loss, we set $\lambda_{\text{asr}} = 1.3$ and $\lambda_{\text{ssim}} = 1.0$, which work well across datasets. We run Algorithm~\ref{algo:search_OAS} for $T = 100$ epochs using the Adam \cite{kingma2014adam} optimizer with a learning rate of 0.01. After training, the learned weights for combining base attacks in each stage are used to generate the final adversarial example. A visualization of this attack combination has been added in the supplementary material.


\begin{figure}[t]
    \centering
    \includegraphics[width=\columnwidth]{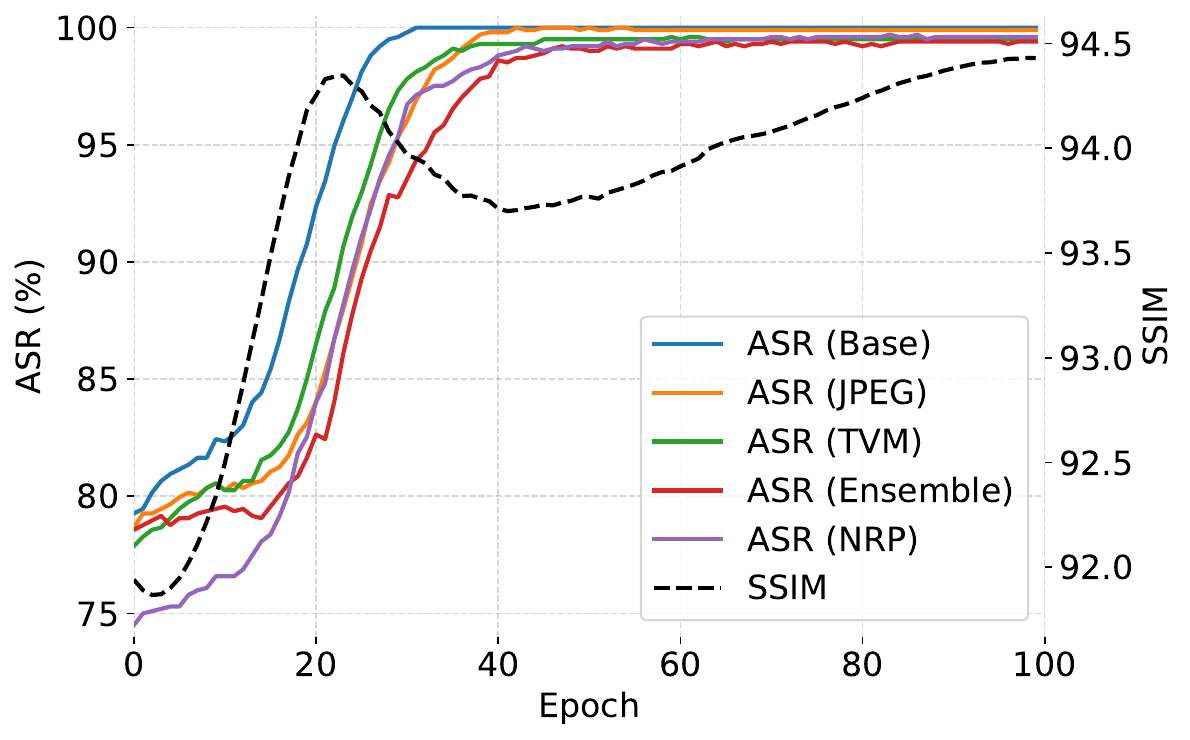}
    \caption{\daash~training progression over epochs. 
    }
    \label{fig:asr_ssim_vs_epoch}
    \vspace{-0.3in}
\end{figure}
\subsection{Effectiveness of \daash~against Robust Models}

\begin{table*}[t]
\centering
\caption{Average attack success rate (ASR↑) and structural similarity (SSIM↑) for cross-model weight transferability.}
\label{tab:transfer_table}

\newcolumntype{P}[1]{>{\centering\arraybackslash}p{#1}}
\resizebox{\textwidth}{!}{%
\renewcommand{\arraystretch}{1}
\small\addtolength{\tabcolsep}{-4pt}
\begin{tabular}{P{1.5cm}|P{2cm}|cccccc|cccccc|cc}
\toprule
      &  \multirow{4}{*}[-0.5ex]{\diagbox[width=1.8cm, height=1.5cm]{\textbf{Trained}}{\textbf{Tested}}}                   & \multicolumn{6}{c|}{\textbf{CIFAR-100}}                 & \multicolumn{6}{c|}{\textbf{CIFAR-10}}                  & \multicolumn{2}{c}{\textbf{ImageNet-1K}} \\ \cmidrule(lr){3-14}\cmidrule(lr){15-16} 
      &                     & \multicolumn{2}{c}{Cui2024 \cite{cui2024decoupled}} 
      & \multicolumn{2}{c}{Wang2023 \cite{wang2023better}} & \multicolumn{2}{c|}{Addepalli22 \cite{addepalli2022efficient}} & \multicolumn{2}{c}{Bartold2024 \cite{bartoldson2024adversarial}} & \multicolumn{2}{c}{Cui2024 \cite{cui2024decoupled}} & \multicolumn{2}{c|}{Wang2023 \cite{wang2023better}} & \multicolumn{2}{c}{Salman20 \cite{salman2020adversarially}} \\ \cmidrule(lr){3-16} 
\textbf{Dataset} &  & ASR & SSIM & ASR & SSIM & ASR & SSIM & ASR & SSIM & ASR & SSIM & ASR & SSIM & ASR & SSIM \\ \midrule
\multirow{3}{*}{CIFAR-100}

 & Cui2024\cite{cui2024decoupled}         & \cellcolor{gray!20}99.77 & \cellcolor{gray!20}94.43 & 99.75 & 94.71 & 99.77 & 94.13 & 99.84 & 93.51 & 99.88 & 93.06 & 99.88 & 92.90 & 99.55 & 95.80 \\
 
 & Wang2023\cite{wang2023better}           & 99.71 & 94.46 & \cellcolor{gray!20}99.67 & \cellcolor{gray!20}94.73 & 99.71 & 94.15 & 99.82 & 93.55 & 99.80 & 93.08 & 99.82 & 92.91 & 99.36 & 95.81 \\
  
 & Addepalli22\cite{addepalli2022efficient}   & 99.80 & 94.45 & 99.75 & 94.69 & \cellcolor{gray!20}99.22 & \cellcolor{gray!20}95.05 & 99.90 & 93.59 & 99.88 & 93.02 & 99.88 & 92.89 & 99.55 & 95.84 \\ \midrule

\multirow{3}{*}{CIFAR-10}
 & Bartold2024\cite{bartoldson2024adversarial}       & 96.85 & 94.73 & 96.84 & 94.92 & 97.04 & 94.48 & \cellcolor{gray!20}97.20 & \cellcolor{gray!20}93.89 & 97.22 & 93.66 & 97.12 & 93.67 & 96.18 & 95.70 \\
 & Cui2024\cite{cui2024decoupled}        & 99.39 & 94.36 & 99.18 & 94.60 & 99.43 & 94.11 & 99.59 & 93.45 & \cellcolor{gray!20}99.61 & \cellcolor{gray!20}93.21 & 99.63 & 93.06 & 98.65 & 95.40 \\
 & Wang2023\cite{wang2023better}          & 98.65 & 94.23 & 98.56 & 94.48 & 98.98 & 93.97 & 98.85 & 93.31 & 98.71 & 93.07 & \cellcolor{gray!20}98.71 & \cellcolor{gray!20}92.91 & 97.89 & 95.36 \\ \midrule
ImageNet
 & Salman20\cite{salman2020adversarially}                 & 97.94 & 94.28 & 97.86 & 94.54 & 98.02 & 93.86 & 98.24 & 93.26 & 98.56 & 92.47 & 98.56 & 92.32 & \cellcolor{gray!20}96.98 & \cellcolor{gray!20}96.52 \\ \bottomrule

\end{tabular}
}
\end{table*}

Table~\ref{tab: main_result} reports the attack success rate (ASR) and perceptual similarity scores of \daash~and seven representative state-of-the-art attacks evaluated against seven recent adversarially trained robust models across CIFAR-10, CIFAR-100, and ImageNet. As shown in Table~\ref{tab: main_result}, \daash~consistently outperforms all baseline attacks in terms of ASR under both the Base (no post-processing) and post-processing defense conditions, across all datasets. This includes outperforming recent attacks explicitly designed to go beyond $\ell_p$-norm constraints. For example, on CIFAR-100, the best-performing baseline is CW with an average ASR of $84\%$, while the most recent perceptual-aligned attack, AdvAD, achieves only $79\%$. In contrast, \daash~achieves an ASR close to $100\%$. Moreover, \daash~yields better imperceptibility scores compared to these dedicated perceptual attacks. Similar trends are observed on CIFAR-10 and ImageNet: \daash~improves over the best baseline by at least $15\%$ ASR on CIFAR-10 and $5\%$ on ImageNet, on average. Its perceptual scores are also consistently better than other attacks specifically designed for perceptual alignment.
For baseline comparison points only, we have obtained results with different parameter configurations and reported the best results. However, for \daash, we used fixed set of hyperparameters for base attacks to ensure fairness.

\vspace{-5px}
\shortsection{Effectiveness of CW}
An interesting finding is that the CW attack, originally developed for $\ell_2$-norm constrained settings, also achieves competitive ASR and perceptual similarity when its hyperparameters are carefully tuned (we tested multiple configurations and report the best result). 
In some cases, CW was also able to outperform DiffPGD, AdvAD, and DiffAttack.
This suggests that some classical attacks, when properly tuned, can potentially still be highly effective in terms of both ASR and imperceptibility.

Although \daash~outperforms CW in terms of ASR, its imperceptibility scores are sometimes slightly lower. Intuitively, since CW is one of the base attacks in \daash, it could be expected to receive higher attention weights during optimization. However, this is not always the case. The reason lies in the mismatch between the objectives: CW uses a hinge-like loss that stops contributing gradients once misclassification is achieved, whereas our meta-loss continues to penalize the confidence of the true class beyond the decision boundary. As a result, after initial misclassification, CW receives lower weights in \daash, which shifts focus to other attacks that still produce informative gradients. 
This continued “pushing” of adversarial examples beyond the boundary enhances robustness of \daash~against additional defenses such as post-processing where CW fails.

\subsection{Effectiveness of \daash~on Black-Box Settings}
Table \ref{tab:transfer-DASH} reports the attack performances of \daash~in black-box scenarios. Additional preliminary results of using black-box attacks (e.g. Sign Flip \cite{chen2020boosting}, Rays \cite{chen2020rays}, Square \cite{andriushchenko2020square}, Sign Hunter \cite{al2020sign}) as \daash~base attacks can be found in the supplementary document (appendix).  

\vspace{-5px}
\shortsection{Adversarial Examples from Surrogate Model} Table \ref{tab:transfer-DASH} shows that the adversarial examples generated with \daash~for one model also perform well for other models. \daash~achieves the strongest transferability, achieving the top average ASR for each model while also maintaining the best perceptual quality (SSIM). For Cui2024 \cite{cui2024decoupled} and Wang2023 \cite{wang2023better} models \daash~outperforms the best baseline DiffAttack with 10\% and 12\% improvement in average ASR (non-diagonal values), keeping the best SSIM score. For Addepalli2022 \cite{addepalli2022efficient} \daash~shows similar average ASR to the best baseline AdvAD while achieving a higher SSIM score. This indicates that \daash~consistently achieves good result in both transferability/imperceptibility scenarios. An extended transferability table is provided in supplementary.


\subsection{Ablation Studies on \daash}

Our \daash~framework consists of two key components: (1) learnable weights for each base attack within every attack cell, and (2) chained attack cells to better escape poor local optima. In this section, we conduct ablation studies to demonstrate the importance of each component.

\vspace{-5px}
\shortsection{Importance of Learnable Weights}
To evaluate the benefit of using learnable weights, we compare the performance of \daash~against a variant where the weights for base attacks are initialized randomly and left untrained. Specifically, the random weights are sampled uniformly in the range $[0,1]$ and used directly without optimization. We repeat this random setup 100 times and report the average results.

Figure~\ref{fig:random_vs_trained} shows that using trained weights significantly improves both ASR and imperceptibility (SSIM) across all types of robust models. Furthermore, Figure~\ref{fig:asr_ssim_vs_epoch} illustrates the ASR and SSIM trends as training progresses over epochs. Both metrics exhibit clear improvement: ASR increases across all settings, and SSIM improves almost monotonically, highlighting the effectiveness of our meta-learning process. These results confirm that the learnable weighting mechanism in \daash~plays a crucial role.

\vspace{-5px}
\shortsection{Importance of Multiple Stages}
To assess the value of chaining multiple stages, we evaluate \daash~with varying numbers of stages. Table~\ref{tab:daash_variant} reports the ASR for different stage configurations. The results show a substantial ASR jump when increasing from one to two stages (from 51.88\% to 99.49\%), suggesting that multi-stage can help escape bad local optima and refine the adversarial example generation. Adding more stages may yield improvements.

\vspace{-5px}
\shortsection{Effectiveness against Unseen Defenses} 
We also examine whether the learned weights in \daash~are transferable across models and datasets. Table~\ref{tab:transfer_table} shows that weights trained on one model or dataset can be applied to unseen models or datasets while still achieving near-optimal results. The variation in average ASR remains within 3\%, and SSIM varies by less than 2. These results demonstrate that \daash~can generalize well across different settings/models.

\section{Conclusion}
In this work, we introduced \daash, a novel differentiable framework for learning optimal compositions of adversarial attacks for both white-box and black-box settings. Unlike traditional methods that heuristically combine attacks, \daash~leverages gradient-based optimization to learn soft weightings over multiple attack modules across stages. \daash~is guided by a meta-loss that jointly captures attack effectiveness and perceptual alignment with human vision. 

\daash~enables the transformation of existing norm-constrained attacks into stronger attacks that also produce perceptually aligned and transferable adversarial examples. Empirical results demonstrate that \daash~ significantly outperforms state-of-the-art perceptually aligned attacks in attack success rate, transferability and imperceptibility scores. \daash~ generalizes well to unseen defenses, establishing itself as a strong and adaptive baseline for future robustness evaluation.
Our results underscore the potential of differentiable search in adversarial example generation and open up new directions for developing more adaptive attack strategies. Future work may explore extending \daash~for attacking generative tasks.

\section*{Acknowledgment}
This material is supported by the National Science Foundation (NSF) under Grant No. 2350363 and 2316399.





%


{
    \small
    \bibliographystyle{ieeenat_fullname}
    \bibliography{daash}
}
\clearpage
\appendix
\section{Appendix}
Below, we provide additional details on the base attacks, defense strategies, and evaluation metrics used in our study (Appendix \ref{sec:attack-defense-background}). We then present the hyperparameters for each base attack and detail how they are selected across different stages of \daash~(Appendix \ref{sec:base-param}). The following sections demonstrate the extension of the \daash~framework to black-box settings (Appendix \ref{sec:bbox-dash}) and examine the performance of \daash~when optimizing with various base attacks and imperceptibility metrics (Appendix \ref{sec:different-base-attack}).

\begin{table*}[!htbp]
\centering
\setlength{\tabcolsep}{2pt}
\renewcommand{\arraystretch}{1}
\caption{Performance of \daash~with Black-Box attacks for CIFAR-10 Dataset.}
\label{tab:black-box}
\resizebox{\textwidth}{!}{
\begin{tabular}{llccccccccc}
\toprule
\multicolumn{2}{c}{} &
\multicolumn{6}{c}{\textbf{ASR (\,$\uparrow$\,)}} &
\multirow{2}{*}{\textbf{SSIM (↑)}} & \multirow{2}{*}{\textbf{LPIPS (↓)}} & 
\multirow{2}{*}{\textbf{FID (↓)}} \\
\cmidrule(lr){3-8} 
\multicolumn{1}{c}{\textbf{Model}} & \multicolumn{1}{c}{\textbf{Attack Method}} &
\multicolumn{1}{c}{\textbf{Base}} &
\multicolumn{1}{c}{\textbf{JPEG}} &
\multicolumn{1}{c}{\textbf{TVM}} &
\multicolumn{1}{c}{\textbf{Ensemble}} &
\multicolumn{1}{c}{\textbf{NRP}} &
\multicolumn{1}{c}{\textbf{Avg.}} &
\multicolumn{1}{c}{} &
\multicolumn{1}{c}{} &
\multicolumn{1}{c}{} \\
\midrule
\multirow{5}{*}{{\begin{tabular}{@{}c@{}}
ResNet-18\\
CIFAR-10
\end{tabular}}}
& Sign Flip   & $\underline{95.27\pm0.92}$ & $26.68\pm1.13$ & $75.04\pm2.27$ & $28.48\pm0.47$ & $26.83\pm0.91$ & $50.46\pm0.64$ & $93.88\pm0.15$ & $\boldsymbol{0.0025\pm0.0002}$ & $57.58\pm1.38$ \\
& Rays   & $94.80\pm0.65$ & $54.28\pm1.29$ & $51.25\pm1.55$ & $43.30\pm0.84$ & $29.43\pm0.99$ & $54.61\pm0.56$ & $94.81\pm0.20$ & $0.0109\pm0.0005$ & $\underline{30.09\pm0.48}$ \\
& Square   & $89.16\pm0.92$ & $\underline{55.43\pm1.36}$ & $\underline{82.03\pm0.64}$ & $\underline{46.21\pm0.85}$ & $29.57\pm1.54$ & $60.48\pm0.80$ & $\underline{95.19\pm0.07}$ & $0.0160\pm0.0005$ & $34.45\pm0.74$ \\
& Sign Hunter   & $85.00\pm1.65$ & $\boldsymbol{58.95\pm1.03}$ & $77.93\pm1.32$ & $\boldsymbol{50.47\pm0.84}$ & $\underline{30.06\pm1.15}$ & $\underline{60.48\pm0.73}$ & $94.81\pm0.15$ & $0.0175\pm0.0002$ & $34.35\pm0.39$ \\
& \cellcolor{gray!15}\textbf{\daash}
& \cellcolor{gray!15}$\boldsymbol{97.79\pm0.34}$
& \cellcolor{gray!15}$50.74\pm0.86$
& \cellcolor{gray!15}$\boldsymbol{88.50\pm0.75}$
& \cellcolor{gray!15}$42.38\pm1.56$
& \cellcolor{gray!15}$\boldsymbol{32.99\pm1.88}$
& \cellcolor{gray!15}$\boldsymbol{62.48\pm0.76}$
& \cellcolor{gray!15}$\boldsymbol{95.85\pm0.07}$
& \cellcolor{gray!15}\underline{$0.0107\pm0.0001$}
& \cellcolor{gray!15}$\boldsymbol{30.10\pm0.43}$ \\
\midrule
				
\multirow{5}{*}{{\begin{tabular}{@{}c@{}}
ViT (base) \\
CIFAR-10
\end{tabular}}}
& Sign Flip   & $94.43\pm0.88$ & $34.98\pm0.90$ & $72.56\pm0.77$ & $30.37\pm1.23$ & $34.04\pm1.30$ & $53.28\pm0.53$ & $92.73\pm0.13$ & $\boldsymbol{0.0034\pm0.0001}$ & $63.57\pm2.24$ \\
& Rays   & $90.51\pm1.10$ & $63.20\pm1.30$ & $61.39\pm1.23$ & $\underline{53.85\pm1.85}$ & $40.18\pm1.40$ & $61.82\pm0.43$ & $95.15\pm0.18$ & $0.0271\pm0.0013$ & $32.15\pm1.12$ \\
& Square   & $93.89\pm0.79$ & $63.54\pm2.41$ & $85.88\pm0.20$ & $46.56\pm1.84$ & $\underline{46.52\pm0.55}$ & $67.28\pm0.92$ & $\underline{96.50\pm0.09}$ & $0.0105\pm0.0004$ & $31.35\pm0.67$ \\
& Sign Hunter   & $\underline{95.86\pm0.51}$ & $\boldsymbol{70.06\pm0.82}$ & $\underline{90.02\pm1.04}$ & $\boldsymbol{59.45\pm1.12}$ & $39.41\pm1.56$ & $\underline{70.96\pm0.54}$ & $96.19\pm0.10$ & $0.0177\pm0.0003$ & \underline{$28.70\pm0.64$} \\
& \cellcolor{gray!15}\textbf{\daash}
& \cellcolor{gray!15}$\boldsymbol{98.20\pm0.36}$
& \cellcolor{gray!15}\underline{$67.49\pm2.28$}
& \cellcolor{gray!15}$\boldsymbol{91.46\pm0.91}$
& \cellcolor{gray!15}$49.64\pm0.76$
& \cellcolor{gray!15}$\boldsymbol{48.73\pm1.27}$
& \cellcolor{gray!15}$\boldsymbol{71.11\pm0.58}$
& \cellcolor{gray!15}$\boldsymbol{97.55\pm0.07}$
& \cellcolor{gray!15}\underline{$0.0081\pm0.0001$}
& \cellcolor{gray!15}$\boldsymbol{24.10\pm0.43}$ \\

\bottomrule
\end{tabular}
}
\end{table*}

\section{Additional Background}\label{sec:attack-defense-background}

\begin{figure}[htbp]
    \centering
    \includegraphics[width=\columnwidth]{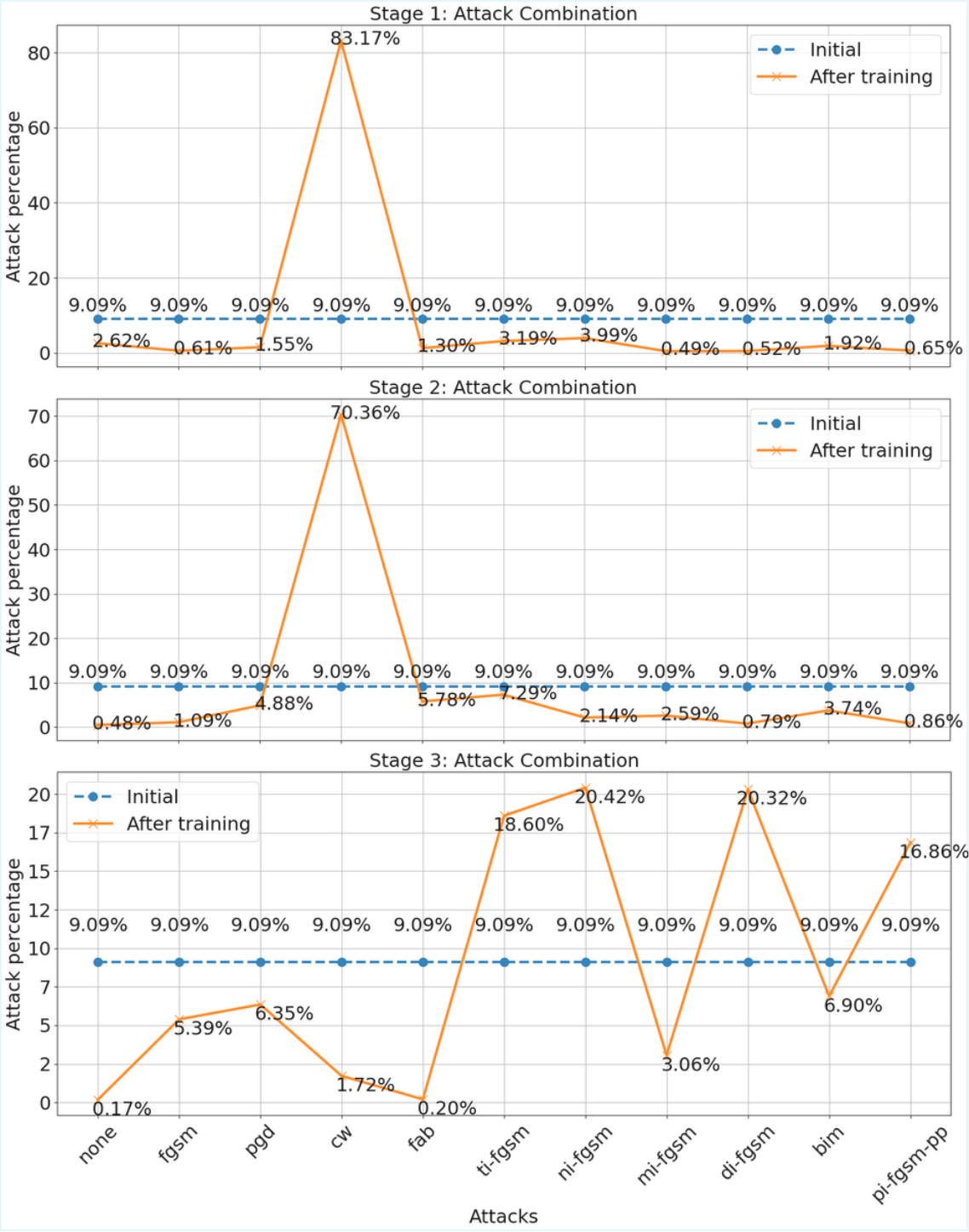}
    \caption{Learned \daash~weights (represented as \% of contribution) for different base attacks across stages. 
    }
    \label{fig:attack_combination}
\end{figure}

\subsection{Norm-Constrained Attacks}
Classical adversarial methods constrain perturbations using $\ell_p$ norms by either enforcing them directly \cite{madry2017towards} or indirectly \cite{carlini2017towards}. 

\textbf{PGD Attack} The PGD attack seeks an adversarial example $\bxa$ such that $ \| \bxa - \bx \|_\infty \leq \epsilon $, where $\epsilon$ denotes the \emph{hard} norm constraint and we denote the norm-constraint as $B_\epsilon$. and $ \bxa $ maximizes a loss $\mathcal{L}$ such as the cross-entropy loss. The update process is iterative. The first iteration is initialized as $ \bxa^{(0)} = x + \delta_0,~~\delta_0 \sim \mathcal{U}(-\epsilon, \epsilon)$ and is later updated in step $t$ as
\[
\bxa = \Pi_{B_\epsilon} \left( \bxa^{(t)} + \alpha \cdot \text{sign}\left( \nabla_x \mathcal{L}(\bxa^{(t)}, y; \theta) \right) \right),
\]
where \( \alpha \) denotes the step size in each iteration, $\Pi$ denotes the projection operation onto the ball $B_\epsilon$, centered at $\bxa$. The attack runs a total of $T$ iterations. FGSM is a single step update: 
\[
\bxa = \Pi_{B_\epsilon} \left( \bx + \epsilon \cdot \text{sign}\left( \nabla_x \mathcal{L}(\bxa^{(t)}, y; \theta) \right) \right).
\] Both the FGSM and PGD attack can be further enhanced, mostly for the purpose of improved transferability to unknown models without white-box access, by providing enhanced inputs to the optimization process \cite{xie2019improving,gao2020patch}. 

\textbf{CW Attack}
The Carlini \& Wagner (CW) \cite{carlini2017towards} attack enforces the norm constraints softly and formulates adversarial example generation as the following optimization without explicit projections:
\begin{equation}\label{eq:cw}
\min_{\boldsymbol{\delta}} \|\boldsymbol{\delta}\|_2^2 + c \cdot \max(0, Z(\bx+\delta)_{y} - \max_{i \neq y} Z(\bx+\boldsymbol{\delta})_i + \kappa)
\end{equation}
where $Z(x)$ represents pre-softmax logits, $c$ balances the two objectives, and $\kappa$ controls confidence margins, which is usually set 0 for common adversarial example generation.

\begin{table*}[t]
\centering
\footnotesize            
\renewcommand{\arraystretch}{1}
\small\addtolength{\tabcolsep}{-2pt}
\caption{Performance comparison of \daash~with varying base attacks for CIFAR-100 Cui2024 \cite{cui2024decoupled} model.}
\label{tab:base-attacks-centered}

\begin{adjustbox}{width=\textwidth,center} 
\begin{tabular}{
  >{\centering\arraybackslash}p{0.25\textwidth}   
  *{9}{c}                                        
}
\toprule
\multicolumn{1}{c}{\multirow{3}{*}{\textbf{Base Attacks}}} &
\multicolumn{6}{c}{\textbf{ASR (\,$\uparrow$\,)}} &
\multicolumn{1}{c}{\multirow{3}{*}{\textbf{SSIM (\,$\uparrow$\,)}}} &
\multicolumn{1}{c}{\multirow{3}{*}{\textbf{LPIPS (\,$\downarrow$\,)}}} &
\multicolumn{1}{c}{\multirow{3}{*}{\textbf{FID (\,$\downarrow$\,)}}} \\
\cmidrule(lr){2-7}
& \makecell{\textbf{W/O}\\ \textbf{Def.}} & \textbf{JPEG} & \textbf{TVM} & \textbf{Ensemble} & \textbf{NRP} & \textbf{Avg} & & & \\
\midrule
\makecell[c]{PGD,\ MI-FGSM,\ CW}
& 99.90 & 97.75 & 99.90 & 95.90 & 97.36 & 98.16 & 90.75 & 0.0191 & 60.31 \\
\midrule
\makecell[c]{NI-FGSM,\ CW,\ FGSM,\\ DI-FGSM,\ None}
& 99.41 & 94.82 & 99.32 & 93.16 & 94.73 & 96.29 & 93.33 & 0.0146 & 49.61 \\
\midrule
\makecell[c]{TI-FGSM,\ PGD,\\ PI-FGSM++,\ CW,\\ MI-FGSM,\ FGSM,\ None}
& 100.00 & 99.90 & 100.00 & 99.51 & 99.71 & 99.82 & 93.90 & 0.0133 & 45.73 \\
\midrule
\makecell[c]{CW,\ BIM,\ NI-FGSM,\ PGD,\\ DI-FGSM,\ TI-FGSM,\\ PI-FGSM++,\ FGSM,\ MI-FGSM}
& 97.75 & 94.34 & 97.46 & 92.77 & 94.82 & 95.43 & 92.35 & 0.0193 & 51.24 \\
\midrule
\makecell[c]{None,\ FGSM,\ PGD,\ CW,\ FAB,\\ TI-FGSM,\ NI-FGSM,\ MI-FGSM,\\ DI-FGSM,\ BIM,\ PI-FGSM++}
& 100.00 & 99.90 & 100.00 & 99.32 & 99.61 & 99.77 & 94.43 & 0.0139 & 41.08 \\
\bottomrule
\end{tabular}
\end{adjustbox}
\end{table*}

\subsection{Perceptually-Informed Attack Design}
Moving beyond norm-constrained attacks, recent methods incorporate perceptual metrics directly into optimization objectives to improve their alignments with human vision. 
\subsubsection{Frequency-Domain Approaches}
Recognizing human visual system characteristics, frequency-based methods constrain perturbations to less perceptible components. Semantic Similarity Attack on High-frequency (SSAH) \cite{luo2022frequency} utilizes discrete wavelet transforms to penalize low-frequency modifications. AdvDrop \cite{duan2021advdrop} applies perturbations in DCT space, dropping high-frequency components that humans cannot perceive. 
\subsubsection{Perceptual Distance Integration}
PerC-AL \cite{zhao2020towards} incorporates perceptual color distance to measure the perturbation magnitude and adds this metric as an additional term in the optimization process:
\begin{equation}
\mathcal{L}_{per} = \sum_{i,j} |\Delta E_{ij}(x, x_{adv})|
\end{equation}
where $\Delta E_{ij}$ computes the International Commission on Illumination (CIE) color differences. 

\subsection{Evaluation Metrics Beyond $\ell_p$ Norms}

\textbf{LPIPS} Learned Perceptual Image Patch Similarity (LPIPS) \cite{zhang2018unreasonable} compares two images $\bx_1$ and $\bx_2$ by measuring the distance between their deep feature representations extracted from a fixed CNN (e.g. AlexNet, VGG16, SqueezeNet) and linearly calibrating those distances with learned weights.
\begin{equation}
d(\bx_1,\bx_2) \;=\;
\sum_{l}\frac{1}{H_{l} W_{l}}
\sum_{h,w}
\left\lVert
\mathbf{w}_{l}\,\odot\!
\bigl(\hat{y}^{\,l}_{1;h,w}-\hat{y}^{\,l}_{2;h,w}\bigr)
\right\rVert_{2}^{2}
\end{equation}
where, \( d(\mathbf{x}_1, \mathbf{x}_2) \) represents the distance between the first image \( \mathbf{x}_1 \) and the second image \( \mathbf{x}_2 \) in the embedding space. The index \( l \) denotes the layer in the network, while \( H_l \) and \( W_l \) are the height and width of the feature map at that layer. The spatial indices \( h \) and \( w \) iterate over the positions in the feature map. \( \hat{y}^l_{1;h,w} \) and \( \hat{y}^l_{2;h,w} \) denote the feature activations of the first and second images, respectively, at layer \( l \) and spatial location \( (h, w) \). \( \mathbf{w}_l \) is a channel-wise weight vector or importance mask applied to the features, and \( \odot \) indicates the element-wise (Hadamard) product. The expression \( \|\cdot\|_2^2 \) denotes the squared \( \ell_2 \) norm, measuring the weighted difference between the two feature representations.

\textbf{FID.} Fréchet Inception Distance (FID)~\cite{heusel2017gans} measures the distributional differences between real and generated images in the feature space of a pretrained network (typically Inception-v3). Let \( \mathcal{X}_{r} \) be the set of real images and \( \mathcal{X}_{g} \) the set of generated images. The FID score is computed as:
\begin{equation}
\operatorname{FID}(\mathcal{X}_{r}, \mathcal{X}_{g}) =
\left\lVert \boldsymbol{\mu}_{r} - \boldsymbol{\mu}_{g} \right\rVert_{2}^{2}
+
\operatorname{Tr} \left(
\boldsymbol{\Sigma}_{r} + \boldsymbol{\Sigma}_{g} - 2\left( \boldsymbol{\Sigma}_{r} \boldsymbol{\Sigma}_{g} \right)^{\tfrac{1}{2}}
\right)
\end{equation}

Here, \( \boldsymbol{\mu}_{r} \) and \( \boldsymbol{\Sigma}_{r} \) are the mean and covariance of the real images’ feature embeddings, while \( \boldsymbol{\mu}_{g} \) and \( \boldsymbol{\Sigma}_{g} \) are those of the generated images. The operator \( \operatorname{Tr}(\cdot) \) denotes the trace of a matrix.
\begin{table*}[htbp]
\centering
\caption{Effectiveness of \daash~with varying loss for the robust CIFAR-100 Cui2024 \cite{cui2024decoupled} model.}
\label{tab:daash_loss_variant}
\renewcommand{\arraystretch}{1}
\small\addtolength{\tabcolsep}{4.5pt}
\begin{tabular}{cccccccccc}
\toprule 
\multirow{2}{*}{\textbf{Loss}} & \multicolumn{6}{c}{\textbf{ASR (↑)}} & 
\multirow{2}{*}{\textbf{SSIM (↑)}} & \multirow{2}{*}{\textbf{LPIPS (↓)}} & 
\multirow{2}{*}{\textbf{FID (↓)}} \\
\cmidrule(lr){2-7}
&  Base & JPEG & TVM & Ensemble & NRP & Avg. & & & \\
\midrule

SSIM & 100.00 & 99.90 & 100.00 & 99.32 & 99.61 & 99.77 & 94.43 & 0.0139 & 41.08 \\
LPIPS & 100.00 & 100.00 & 100.00 & 99.41 & 99.80 & 99.84 & 92.34 & 0.0155 & 45.89 \\

\bottomrule
\end{tabular}
\end{table*}

\subsection{Defense Strategies}
Below, we provide a brief overview of the defense mechanisms evaluated against our \daash~framework, categorized into robust models and post-processing defenses.

\subsubsection{Robust Models}
Robust models are trained to be inherently resilient to adversarial examples. \citet{madry2018towards} first proposed \emph{Adversarial Training}, which injects adversarially perturbed inputs (with correct labels), typically generated via PGD attacks, into the training process to enhance robustness. TRADES~\cite{zhang2019theoretically} improves upon adversarial training by replacing the standard cross-entropy loss with a Kullback–Leibler (KL) divergence-based loss that better balances accuracy and robustness. 

Subsequent methods further refine these ideas, building on adversarial training and TRADES. For example, Decoupled Adversarial Learning~\cite{cui2024decoupled}, Better Adversarial Training~\cite{wang2023better}, Adversarial Weight Perturbation~\cite{bartoldson2024adversarial}, and Efficient Robust Training~\cite{addepalli2022efficient} introduce modifications to improve training efficiency, robustness, or generalization.
\begin{table}[t]
\centering
\setlength{\tabcolsep}{2pt}
\renewcommand{\arraystretch}{1.3}
\small\caption{Extended Transferability Comparison on CIFAR-100 Dataset. S represents Surrogate models and T represents Target models.}
\label{tab:transfer-full}
\resizebox{\linewidth}{!}{%
\begin{tabular}{c l c c c c | c c}
\toprule
\multicolumn{1}{c}{\diagbox{\textbf{S (\textdownarrow)}}{\textbf{T (\textrightarrow)}}} & \multicolumn{1}{l}{\textbf{Attack}} & \multicolumn{1}{c}{\textbf{\cite{cui2024decoupled}}} & \multicolumn{1}{c}{\textbf{\cite{wang2023better}}} & \multicolumn{1}{c}{\textbf{\cite{addepalli2022efficient}}} & \multicolumn{1}{c}{\textbf{\cite{debenedetti2023light}}} & \multicolumn{1}{c}{\textbf{Avg.(\textuparrow)}} & \multicolumn{1}{c}{\textbf{SSIM (\textuparrow)}} \\
\midrule
\multirow{6}{*}{\textbf{\shortstack{\cite{cui2024decoupled}}}} & PI-FGSM++ & \cellcolor{gray!15}62.20 & 59.25 & 60.01 & 59.71 & 59.66 & 91.12 \\
& CW & \cellcolor{gray!15}84.14 & 51.39 & 56.48 & 52.44 & 53.44 & \underline{94.38} \\
& AutoAttack & \cellcolor{gray!15}76.77 & 65.98 & 64.26 & 62.11 & 64.12 & 92.09 \\
& DiffAttack & \cellcolor{gray!15}77.24 & 74.76 & 74.72 & 57.10 & 68.86 & 91.16 \\
& AdvAD & \cellcolor{gray!15}79.14 & 73.84 & 72.70 & 64.32 & \underline{70.29} & 83.18 \\
& \daash & \cellcolor{gray!15}99.77 & 89.55 & 79.69 & 70.61 & \textbf{79.95} & \textbf{94.43} \\
\midrule
\multirow{6}{*}{\textbf{\shortstack{\cite{wang2023better}}}} & PI-FGSM++ & 59.96 & \cellcolor{gray!15}63.55 & 61.43 & 59.82 & 60.40 & 91.14 \\
& CW & 52.73 & \cellcolor{gray!15}83.57 & 56.78 & 52.36 & 53.96 & \underline{94.55} \\
& AutoAttack & 63.98 & \cellcolor{gray!15}78.34 & 64.26 & 61.91 & 63.38 & 91.25 \\
& DiffAttack & 75.38 & \cellcolor{gray!15}76.53 & 74.70 & 57.50 & 69.19 & 90.90 \\
& AdvAD & 72.98 & \cellcolor{gray!15}77.42 & 72.24 & 66.40 & \underline{70.54} & 83.07 \\
& \daash & 94.53 & \cellcolor{gray!15}99.67 & 80.64 & 73.20 & \textbf{82.79} & \textbf{94.73} \\
\midrule
\multirow{6}{*}{\textbf{\shortstack{\cite{addepalli2022efficient}}}} & PI-FGSM++ & 53.24 & 53.77 & \cellcolor{gray!15}68.47 & 42.99 & 50.00 & 91.14 \\
& CW & 37.32 & 37.73 & \cellcolor{gray!15}84.18 & 57.46 & 44.17 & \underline{94.96} \\
& AutoAttack & 52.13 & 54.08 & \cellcolor{gray!15}77.90 & 58.22 & 54.81 & 93.03 \\
& DiffAttack & 72.44 & 54.30 & \cellcolor{gray!15}81.35 & 54.70 & 60.48 & 91.26 \\
& AdvAD & 64.58 & 64.82 & \cellcolor{gray!15}81.60 & 61.70 & \textbf{63.70} & 83.92 \\
& \daash & 63.50 & 63.09 & \cellcolor{gray!15}99.22 & 62.83 & \underline{63.14} & \textbf{95.05} \\
\midrule
\multirow{6}{*}{\textbf{\shortstack{\cite{debenedetti2023light}}}} & PI-FGSM++ & 54.86 & 55.25 & 60.29 & \cellcolor{gray!15}68.03 & 56.80 & 91.30 \\
& CW & 41.56 & 41.58 & 48.03 & \cellcolor{gray!15}79.90 & 43.72 & \underline{94.30} \\
& AutoAttack & 52.32 & 53.57 & 60.53 & \cellcolor{gray!15}76.25 & 55.47 & 93.22 \\
& DiffAttack & 61.46 & 63.20 & 67.22 & \cellcolor{gray!15}72.24 & \underline{63.96} & 93.87 \\
& AdvAD & 60.42 & 61.04 & 66.54 & \cellcolor{gray!15}74.52 & 62.67 & 87.28 \\
& \daash & 73.59 & 74.47 & 78.87 & \cellcolor{gray!15}99.57 & \textbf{75.64} & \textbf{94.57} \\
\bottomrule
\end{tabular}
}
\vspace{-0.2in}
\end{table}
\subsubsection{Post-Processing Defenses}
Post-processing defenses aim to increase the robustness of a given model (standard or robust) by transforming or purifying inputs at inference time.

JPEG Compression~\cite{guo2017countering} applies a lossy compress–decompress operation before classification, which removes high-frequency components where adversarial noise typically resides. Total Variation Minimization (TVM)~\cite{wang2018adversarial} similarly aims to suppress high-frequency perturbations through optimization-based denoising.

Neural Representation Purifier (NRP)~\cite{naseer2020self} is a more advanced approach that learns a purification network. Given a potentially adversarial image, this network iteratively adjusts pixels until a frozen reference model confidently recognizes the input as a clean example. NRP has demonstrated strong empirical performance and is widely adopted in recent adversarial defense research.

\section{Base Attack Parameters and Their Selection Across Stages}\label{sec:base-param}
For the base attacks, we use Torchattacks~\cite{kim2020torchattacks}, a PyTorch-based library that provides implementations of various adversarial attacks. The parameters for each attack used in our experiments are detailed below.

For $\ell_{\infty}$-norm attacks, we set the perturbation budget to $\epsilon = \frac{7}{255}$. For different variants of FGSM evaluated in this paper, we use a momentum factor of 1. In TI-FGSM~\cite{dong2019evading} and DI-FGSM~\cite{xie2019improving}, we set the \texttt{resize\_rate} to 0.9, \texttt{diversity\_prob} to 0.7, and \texttt{random\_start} to \texttt{False}. In PI-FGSM++~\cite{gao2020patch}, we set the probability of using diverse inputs to 0.7 and the \texttt{project\_factor} to 0.8.

For the Fast Adaptive Boundary (FAB) attack~\cite{croce2020minimally}, we set $\alpha_{\text{max}} = 0.1$, overshooting factor $\eta = 1.05$, and the backward step parameter $\beta = 0.9$. For BIM~\cite{kurakin2016adversarial} and PGD~\cite{madry2017towards}, we use a step size of $\alpha = 0.01$ and number of iterations $T = 10$.

For the CW attack~\cite{carlini2017towards}, we set the confidence parameter $c = 10$, margin parameter $\kappa = 0$, and optimize for 100 steps using a learning rate of $0.01$.

\subsection{Base Attack Selection Across Stages}\label{sec:base-select}
\daash~learns the parameter values during training that are used to combine multiple base attacks at each stage of the adversarial example generation process. In Figure~\ref{fig:attack_combination}, we visualize the selected attack profiles for a robust model trained on CIFAR-100~\cite{cui2024decoupled}. The blue dotted line represents the initial combination of base attacks, while the orange line shows the learned combination after training.

Initially, the CW attack is selected more heavily due to its strong misclassification and better perceptual preservation capability. However, as training progresses, \daash~shifts weight away from CW, since it no longer contributes to the meta-loss, largely due to the zero-gradient problem introduced by the hinge loss (see Eq.~\eqref{eq:cw}) with $\kappa = 0$. As a result, our attack strategy continues to optimize for stronger and more confident adversarial examples, enhancing its transferability and effectiveness against various unseen post-processing defenses. This comes at the cost of slightly worse perceptual similarity scores, due to the increased distortion required to maintain high confidence.

\section{\daash~on Black-Box Attacks}\label{sec:bbox-dash}
For measuring the performance of \daash~with the Black-Box attack, we have taken 2 query-based attacks, Sign-Flip \cite{chen2020boosting} and Rays \cite{chen2020rays}, and 2 decision-based attacks, Sign-Hunter \cite{al2020sign} and Square attack \cite{andriushchenko2020square}, and combined them using \daash. They are the current state-of-the-art black-box attacks. We have measured the metrics for 5 disjoint sets of 1,000 images and reported the mean$\pm$std. Table \ref{tab:black-box} shows the performance comparison of \daash~with other SOTA black-box attacks for 2 different models, ResNet-18 \cite{he2016deep}, and Vision Transformer \cite{dosovitskiy2020image}, trained on the CIFAR-10 dataset. The table shows that \daash~achieves the highest average attack success rate for the models, exceeding the strongest baselines by almost 2\%, and 1\%. It also maintains the best imperceptibility with the highest SSIM and FID scores. This result implies that \daash~can also be extended to black-box attacks.

\section{\daash~with Varying Base Attacks}\label{sec:different-base-attack}
Table \ref{tab:base-attacks-centered} shows the \daash~performance with varying base attacks. In all the combinations \daash~shows an average attack success rate over 95\% and SSIM over 90\%, which implies that \daash~is compatible with diverse base attacks and consistently achieves improved performance. 


In our main experiment, we used a meta-loss for optimizing attack weights in \daash, balancing attack success rate (ASR) and imperceptibility. Focusing on metrics aligned with the human visual system, we primarily report results using SSIM [$
\mathcal{L}_{\text{total}} = \lambda_{\text{asr}} \cdot f_y(\bxa) + \lambda_{\text{ssim}} \cdot (1 - \text{SSIM}(\bx, \bxa))
$], but we also experimented with LPIPS 
[$
\mathcal{L}_{\text{total}} = \lambda_{\text{asr}} \cdot f_y(\bxa) + \lambda_{\text{lpips}} \cdot \text{LPIPS}(\bx, \bxa)
$] and achieved similar output. Table \ref{tab:daash_loss_variant} demonstrates that \daash~ maintains a high ASR with minimal perturbation when optimizing with both SSIM and LPIPS. Specifically, the ASR varies by less than 0.1\% and the SSIM score varies by only 0.02 units (assuming the 0 to 1 range), confirming the robustness of the framework regardless of the chosen perceptibility metric.





\section{Extended Transfer Result}
Besides the transfer results among multiple versions of Wide-ResNets, we also evaluated strict architectural transferability using XCiT-M12 \cite{debenedetti2023light} (a Transformer), as reported in Table~\ref{tab:transfer-full}. The result shows that the adversarial examples generated with \daash~for one model also perform well for other models. \daash~achieves the strongest transferability, achieving the top average ASR for each model while also maintaining the best perceptual quality (SSIM). For Cui2024 \cite{cui2024decoupled}, and Wang2023 \cite{wang2023better} models \daash~outperforms the best baseline AdvAD with almost 10\% and 12\% improvement in average ASR, keeping the best SSIM score. For Addepalli2022 \cite{addepalli2022efficient} \daash~shows similar performance to the best baseline AdvAD while achieving a higher SSIM score. For Debenedetti2023 \cite{debenedetti2023light}, \daash~outperforms the best baseline DiffAttack with almost 12\% improvement in average ASR while maintaining the highest SSIM score. This indicates that \daash~consistently achieves good result in both transferability and imperceptibility scenarios.

\end{document}